\documentclass[lettersize,journal]{IEEEtran}
\usepackage{amsmath,amsfonts}
\usepackage{array}
\usepackage[caption=false,font=normalsize,labelfont=sf,textfont=sf]{subfig}
\usepackage{textcomp}
\usepackage{stfloats}
\usepackage{url}
\usepackage{verbatim}
\usepackage{graphicx}
\usepackage{cite}

\usepackage{booktabs}
\usepackage{multirow}
\usepackage{tabularx}
\usepackage{adjustbox}
\usepackage{amssymb}
\usepackage{algorithm}
\usepackage{algpseudocode} 

\begin{document}

\title{Lightweight and Scalable Transfer Learning Framework for Load Disaggregation}

\author{L.E. Garcia-Marrero, G. Petrone, and E. Monmasson  
\thanks{L.E. Garcia-Marrero and E. Monmasson are with the Systèmes et Applications des Technologies de l’Information et de l’Energie laboratory, Cergy-Pontoise.}

\thanks{G. Petrone and also L.E. Garcia-Marrero are with the Dipartimento di Ingegneria dell’Informazione ed Elettrica e Matematica Applicata/DIEM, University of Salerno, 84084 Fisciano.}}



\maketitle

\begin{abstract}
Non-Intrusive Load Monitoring (NILM) aims to estimate appliance-level consumption from aggregate electrical signals recorded at a single measurement point. In recent years, the field has increasingly adopted deep learning approaches; however, cross-domain generalization remains a persistent challenge due to variations in appliance characteristics, usage patterns, and background loads across homes. Transfer learning provides a practical paradigm to adapt models with limited target data. However, existing methods often assume a fixed appliance set, lack flexibility for evolving real-world deployments, remain unsuitable for edge devices, or scale poorly for real-time operation. This paper proposes RefQuery, a scalable multi-appliance, multi-task NILM framework that conditions disaggregation on compact appliance fingerprints, allowing one shared model to serve many appliances without a fixed output set. RefQuery keeps a pretrained disaggregation network fully frozen and adapts to a target home by learning only a per-appliance embedding during a lightweight backpropagation stage. Experiments on three public datasets demonstrate that RefQuery delivers a strong accuracy-efficiency trade-off against single-appliance and multi-appliance baselines, including modern Transformer-based methods. These results support RefQuery as a practical path toward scalable, real-time NILM on resource-constrained edge devices.
\end{abstract}


\begin{center}
\small
\textsc{Nomenclature}
\end{center}

\vspace{0.2em}

\noindent
\begin{minipage}[t]{0.48\textwidth}
\raggedright

\textit{Sets and Indices}\par
\vspace{0.15em}
\hspace*{1em}%
\begin{tabular}[t]{@{}ll@{}}
$b$ & Building index \\
$k$ & Appliance index \\
$t$ & Time index \\
$\mathcal{H}$ & Set of buildings \\
$\mathcal{K}$ & Set of target appliances \\
\end{tabular}

\vspace{0.65em}

\textit{Parameters}\par
\vspace{0.15em}
\hspace*{1em}%
\begin{tabular}[t]{@{}ll@{}}
$L$ & Window length \\
$E$ & Embedding dimension \\
$H$ & Width of shared fully connected layers \\
$\theta$ & Parameters of the feature extractor \\
$\phi$ & Parameters of the prediction head \\
$b_{\mathrm{off}}$ & Fixed OFF bias in z-space \\
\end{tabular}

\end{minipage}
\hfill
\begin{minipage}[t]{0.48\textwidth}
\raggedright

\textit{Variables}\par
\vspace{0.15em}
\hspace*{1em}%
\begin{tabular}[t]{@{}ll@{}}
$x_b(t)$ & Aggregate mains signal of building $b$ \\
$y_b^{(k)}(t)$ & Appliance $k$ power signal \\
$q \in \mathbb{R}^{L}$ & Query mains window \\
$r^{(k)} \in \mathbb{R}^{L}$ & Reference window for appliance $k$ \\
$\mathbf{e}_r,\mathbf{e}_q \in \mathbb{R}^{E}$ & Reference and query embeddings \\
$e_r^{(k)}$ & Reference embedding for appliance $k$ \\
$\tilde{e}_r^{(k)}$ & Learned target-domain reference embedding \\
$\hat{y}^{(k)}$ & Predicted appliance power \\
$\hat{s}^{(k)}$ & Predicted ON/OFF state \\
$\hat{y}_z$ & Predicted z-normalized power \\
$\mathcal{L}$ & Total loss \\
$\mathcal{L}_{\mathrm{MSE}}$ & Power loss \\
$\mathcal{L}_{\mathrm{BCE}}$ & State loss \\
\end{tabular}

\end{minipage}

\vspace{0.4em}
\section{Introduction}
\label{sec:intro}

\IEEEPARstart{N}{on}-Intrusive Load Monitoring (NILM), also known as energy disaggregation, aims to extract appliance-level consumption data from aggregate electrical signals recorded at a single measurement point \cite{Murray2019}. For NILM to be practical, it must operate under constraints such as the low sampling rates and limited electrical measurements typically provided by low-cost commercial smart meters \cite{GarciaMarrero2025}. Additionally, it must be capable of running on edge devices due to growing concerns about data privacy and transmission costs \cite{GarciaMarrero2025}. From this standpoint, all target-domain computing should occur locally on the device to preserve user privacy and reduce operational complexity \cite{Tanoni2024}.

The release of large-scale household energy usage datasets, such as REDD \cite{kolter2011redd}, UK-DALE \cite{Kelly2015}, and REFIT \cite{Murray2017}, has supported the growing use of deep learning (DL) approaches in energy disaggregation. DL models can learn discriminative representations from raw data and often improve disaggregation accuracy in scenarios involving complex load mixtures. These improvements have expanded the range of appliances that can be effectively disaggregated, including complex multi-state devices that remain difficult for classical models to handle \cite{Kong2020}.

Despite the accuracy of recent DL-based NILM, several state-of-the-art proposals are not designed under the constraints of edge deployment, where memory and compute are limited. In particular, some accuracy gains are achieved by adopting architectures that increase parameter count and inference-time computation. For example, SAMNet employs a multi-task design with specialized expert and attention mechanisms to exploit interactions between state detection and energy estimation, increasing architectural complexity \cite{Liu2022}. MATNilm, similarly, introduces a multi-appliance-task framework with additional correlation modeling across appliances, adding components beyond a simple shared backbone \cite{Xiong2024}. In addition, PSSR-Net adopts a deep feature extractor (ResNet50) as its backbone, which is costly for on-device inference and makes any local fine-tuning more demanding \cite{Yang2025}.

 Moreover, most approaches adopt a single-appliance paradigm (one model per appliance), growing storage linearly with the number of targets. Multi-appliance models, in contrast, reduce redundancy by sharing early layers and either producing outputs for a predetermined appliance set \cite{Li2023}, or attaching appliance-specific sub-networks \cite{Sun2025}, which scales poorly as the number of appliances increases. A more suitable direction for edge deployment is a single model that can address many appliances incrementally, specified through a compact fingerprint. Recent steps in this direction include conditioned multi-target disaggregation \cite{Garca2024} and query-based models using appliance embeddings \cite{Yang2025_QueryNILM}; however, this research direction is still in a preliminary phase.

On the other hand, generalization remains a primary barrier, since models trained in a source domain often degrade when deployed in a target home due to differences in appliance characteristics, usage patterns, and background loads \cite{Murray2019, GarciaMarrero2026}. A first family of approaches attempts to improve generalization without needing target-domain data by enriching the training distribution. In \cite{Rafiq2021}, a data-augmentation strategy is proposed to synthesize diverse appliance activations for training. In \cite{Luo2024}, a metric-based meta-learning framework is proposed to improve generalizability across domains and label spaces without target-domain retraining. In \cite{Liu2024}, a self-supervised feature-learning method is proposed to learn representations for appliance recognition under label scarcity. However, these approaches rely on source-side diversity to anticipate deployment conditions, which is insufficient to guarantee robust performance across unseen homes. 

A second family of methods addresses generalization using only target-domain aggregate data to reduce domain shift. In \cite{Chen2023}, self-supervised learning is used to pretrain a feature extractor on unlabeled target aggregate signals, and the pretrained weights are then used to initialize the downstream disaggregation model. In \cite{Lin2022}, domain adaptation is proposed to align source and target distributions through additional objectives computed on unlabeled target aggregates. While these approaches reduce the need for labeled appliance power, they require collecting and processing substantial aggregated data, which is challenging under tight resource constraints. Moreover, adapting solely from aggregated labels does not fully resolve mismatches that arise from variations in appliance instances across different homes.

A more practical approach to address the generalization challenge is transfer learning (TL), which adapts a model trained in a source domain to a target domain using a small amount of target-specific data. In \cite{DIncecco2020}, the authors propose TL by fine-tuning the final dense layers while keeping the backbone frozen. In \cite{Wang2022}, a meta-learning-based approach and an ensemble-learning-based approach are proposed, followed by target adaptation. In \cite{Li2023}, cross-domain adaptation is performed via supervised retraining of the top dense layers and subsequent fine-tuning of the shared CNN layers.
In \cite{Sun2025}, transfer learning is performed by freezing the shared module and fine-tuning the appliance-specific layers. However, practical transfer learning must enable lightweight adaptation with limited data, operate under hardware constraints, and support scalable, real-time deployment.

This paper presents RefQuery, a scalable multi-appliance multi-task NILM framework that supports incremental adaptation of appliances by learning a lightweight per-appliance embedding during a short backpropagation-based adaptation stage, while keeping a pretrained disaggregation network completely frozen. This clean separation between a stable global model and appliance-specific embeddings enables plug-and-play expansion, simplifies deployment and versioning, and restricts adaptation to a compact, well-defined set of parameters rather than network fine-tuning. The contributions of this work are summarized as follows:

\begin{itemize}
    \item Development of an appliance-conditioned NILM framework that enables multi-appliance disaggregation with a single shared model by conditioning inference on compact appliance embeddings, improving deployment practicality on constrained edge devices.

    \item Introduction of a novel source-domain training strategy and a novel target-domain adaptation mechanism that allows appliances to be disaggregated without updating network parameters through a lightweight adaptation stage.

    \item Validation of the proposed solution through extensive experiments on cross-domain transfer learning, limited-data adaptation, and computational scalability, showing that RefQuery achieves competitive accuracy while substantially reducing storage and adaptation costs compared with state-of-the-art baselines, supporting its suitability for real-time edge deployment.
\end{itemize}

The remainder of this paper is organized as follows. Section \ref{sec:method} describes the proposed RefQuery framework. Section \ref{sec:arch} presents the model architecture. Section \ref{sec:experiments} outlines the experimental setup, reports the results, and provides the corresponding analysis, and section \ref{sec:conclusion} concludes the paper.

\section{Method}
\label{sec:method}

\subsection{Framework overview}
\label{subsec:framework_overview}

This paper presents RefQuery, a multi-task, appliance-conditioned NILM framework. RefQuery estimates both appliance-level power and activation state from aggregate mains by combining a reference appliance embedding with a query mains embedding within a shared model. For each appliance $k$, assume that there exists a fixed reference window $r^{(k)} \in \mathbb{R}^{L}$ of length $L$ that contains a representative activation pattern of the device. Given a query mains window $q \in \mathbb{R}^{L}$ of length $L$, the reference and query signals are encoded by a shared feature extractor $f_{\theta}(\cdot)$ producing embeddings of equal dimensionality $E$,
\begin{equation}
e_r^{(k)} = f_{\theta}\!\left(r^{(k)}\right) \in \mathbb{R}^{E}, 
\qquad 
e_q = f_{\theta}(q) \in \mathbb{R}^{E},
\end{equation}
These embeddings are combined using a prediction head $h_{\phi}(\cdot,\cdot)$ to jointly estimate the appliance's power and on/off state:
\begin{equation}
(\hat{y}^{(k)}, \hat{s}^{(k)}) = h_{\phi}\!\left(e_r^{(k)}, e_q\right),
\end{equation}
where $\hat{y}^{(k)}$ denotes the estimated appliance power and $\hat{s}^{(k)}$ the estimated on/off (activation) state. Conditioning on $e_r^{(k)}$ allows a single backbone to be reused across appliances by changing only the reference embedding, avoiding per-appliance models and reducing the rigidity of architectures tied to a fixed appliance set, which is beneficial for constrained edge deployment. 

\begin{figure*}[!t]
    \centering
    \includegraphics[width=1\linewidth]{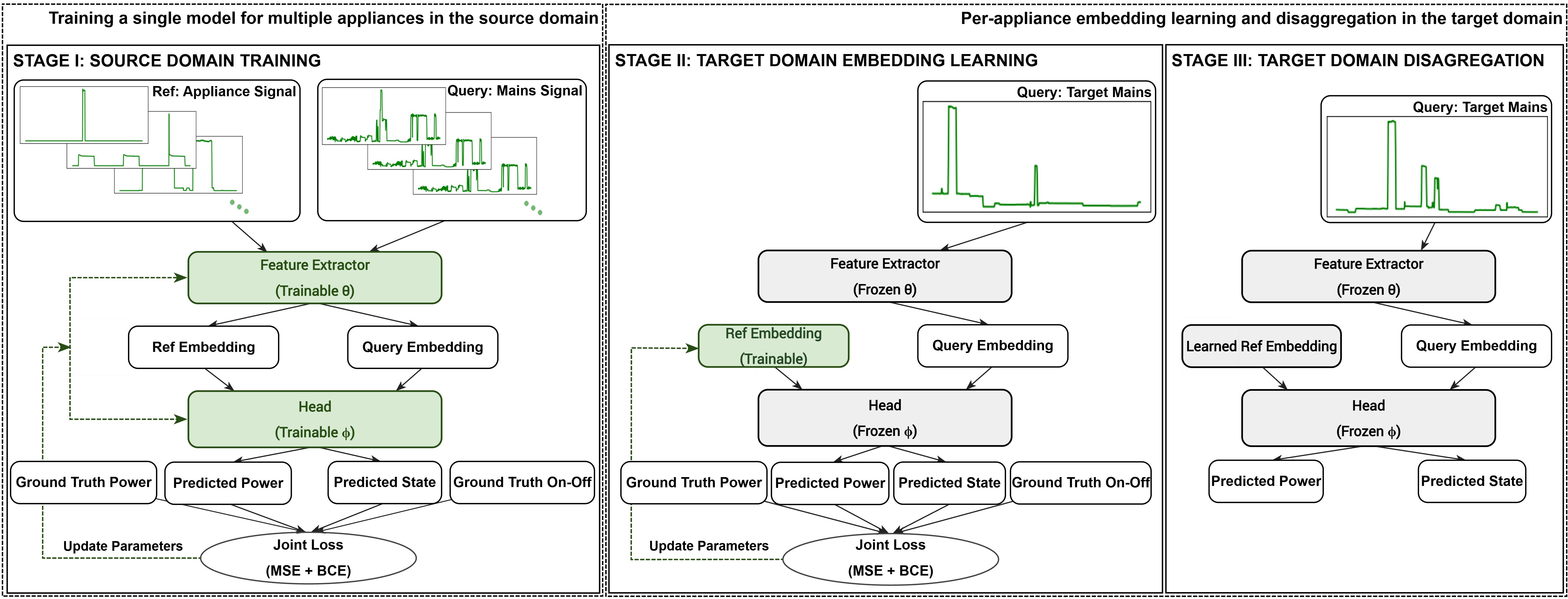}
    \caption{Three-stage RefQuery framework. Stage I: the model is trained on multiple source buildings using several appliances per building with joint power (MSE) and state (BCE) supervision; Stage II: in the target domain, the model is frozen and a target-specific reference embedding is learned per appliance using joint power (MSE) and state (BCE) supervision; Stage III: target mains is disaggregated into per-appliance state and power using the learned embeddings.}
    \label{fig:refquery_flow}
\end{figure*}

RefQuery is first trained in a source domain using fully labelled data, so that the shared feature extractor $f_{\theta}(\cdot)$ and head $h_{\phi}(\cdot,\cdot)$ learn a reusable, appliance-conditioned disaggregation function. At deployment in an unseen household, RefQuery performs lightweight adaptation by freezing $f_{\theta}$ and $h_{\phi}$ and learning only the appliance reference embedding. For each appliance $k$, we initialize $e_r^{(k)} \in \mathbb{R}^{E}$ at random and optimize it to minimize prediction error on labeled target-domain data. The same procedure is repeated each time a new appliance is added. Since the reference embedding has a dimensionality of $E$, adaptation in the target domain requires optimizing only $E$ parameters per appliance, rather than fine-tuning the full network, which substantially reduces computation and memory requirements, making personalization feasible on constrained edge devices. Figure \ref{fig:refquery_flow} illustrates the overall structure of the RefQuery framework, and the following sections provide a detailed explanation of each stage.

\subsection{Stage I: Training on the Source Domain}
\label{subsec:stage1}
In Stage~I, RefQuery is trained in a source domain where both aggregate mains and sub-metered appliance signals are available. We denote by $x_b(t)$ the aggregate mains signal of building $b$, and by $y_{b}^{(k)}(t)$ the power consumption of appliance $k$ in the same building. For simplicity, we assume that (i) all target appliances $k\in\mathcal{K}$ are present in every building $b\in\mathcal{H}$, and (ii) $x_b(t)$ and $y_{b}^{(k)}(t)$ are time-aligned and sampled at the desired sampling period $s$; otherwise, additional preprocessing (e.g., resampling and alignment) must be applied before generating the training set.

For each building $b$ and appliance $k$, we generate training quadruples according to Algorithm~\ref{alg:train_gen_refquery}. Briefly, we first construct a reference bank $\mathcal{R}_{b,k}$ by collecting candidate reference windows for appliance $k$ in building $b$. Specifically, for each ON-interval of $y_{b}^{(k)}(t)$ we extract a length-$L$ window centered at the midpoint of that interval. This yields a set of segments that capture the representative active behavior of appliance $k$ in building $b$. Here, it is convenient to set $L$ large enough to capture a complete activation pattern for each target appliance, since $L$ is a fixed window length shared across appliances. Then, for every time index $t$, a query window from mains $q = x_b[t:t{+}L]$ is paired with a reference window sampled from $\mathcal{R}_{b,k}$. We adopt a Seq2Point formulation \cite{DIncecco2020}, where supervision is defined at the window center $t+p$ with $p=\lfloor L/2\rfloor$. The regression label is then the instantaneous appliance power $y^{P}=y_{b}^{(k)}[t{+}p]$, while the classification label $y^{ON}\in\{0,1\}$ indicates whether appliance $k$ is ON at $t{+}p$.

\begin{algorithm}[t]
\caption{Training-set generation (Stage I)}
\label{alg:train_gen_refquery}
\begin{algorithmic}[1]
\Require Window length $L$; sampling period $s$; building set $\mathcal{H}$; target appliance set $\mathcal{K}$; for each building $b\in\mathcal{H}$: mains signal $x_b$ and appliance signals $\{y_{b}^{(k)}\}_{k\in\mathcal{K}}$; ON-intervals $\{\mathcal{I}_{b,k}\}_{k\in\mathcal{K}}$.

\Ensure Training sets $\mathcal{Q}$ (queries), $\mathcal{R}$ (references), $\mathcal{Y}^{P}$ (power), $\mathcal{Y}^{ON}$ (state), fitted z-normalizers.

\State $p \gets \lfloor L/2 \rfloor$
\State Initialize $\mathcal{Q},\mathcal{R},\mathcal{Y}^{P},\mathcal{Y}^{ON} \gets [\ ]$
\For{each building $b\in\mathcal{H}$}
  \For{each appliance $k\in\mathcal{K}$}
    \State $N \gets |y_{b}^{(k)}|$
    \State \textbf{Reference bank} $\mathcal{R}_{b,k} \gets [\ ]$
    \For{each ON-interval $[s,e)$ in $\mathcal{I}_{b,k}$}
      \State $c \gets \lfloor (s+e)/2 \rfloor$;\quad $t_0 \gets c - \lfloor L/2 \rfloor$
      \If{$0 \le t_0$ \textbf{and} $t_0+L \le N$}
        \State append $y_{b}^{(k)}[t_0:t_0{+}L]$ to $\mathcal{R}_{b,k}$
      \EndIf
    \EndFor
    \For{$t=0$ to $N-L$}
      \State append $x_b[t:t{+}L]$ to $\mathcal{Q}$
      \State append $\textsc{Sample}(\mathcal{R}_{b,k})$ to $\mathcal{R}$
      \State append $y_{b}^{(k)}[t{+}p]$ to $\mathcal{Y}^{P}$
      \If{$y_{b}^{(k)}$ is ON at $t{+}p$}
        \State append $1$ to $\mathcal{Y}^{ON}$
      \Else
        \State append $0$ to $\mathcal{Y}^{ON}$
      \EndIf
    \EndFor
  \EndFor
\EndFor
\State fit z-normalizers on $\mathcal{R},\mathcal{Q},\mathcal{Y}^{P}$; normalize $\mathcal{R},\mathcal{Q},\mathcal{Y}^{P}$
\State \Return $\mathcal{Q},\mathcal{R},\mathcal{Y}^{P},\mathcal{Y}^{ON}$
\end{algorithmic}
\end{algorithm}

\subsubsection{ON Intervals Construction}
\label{subsubsec:on_intervals}

For each appliance signal $y_{b}^{(k)}$, we construct the ON-interval set $\mathcal{I}_{b,k}$ used in Algorithm~\ref{alg:train_gen_refquery} by computing a binary activation mask and then extracting contiguous ON segments. We first create a preliminary binary mask by marking a sample as ON whenever the appliance power exceeds a threshold (\texttt{20W}). This yields a raw activation mask that may still include short spurious bursts or brief OFF gaps due to noise and transient fluctuations. To ensure physically consistent activations, we apply two duration-based post-processing rules:

(i) ON runs shorter than \texttt{60s} are removed (set to OFF), and

(ii) OFF gaps shorter than \texttt{300s} are bridged (set to ON) when they are surrounded by ON samples on both sides.

Finally, the cleaned mask is converted into ON-intervals by collecting all maximal contiguous segments where the mask is equal to 1. 

\subsubsection{Training Loss}
\label{subsubsec:training_loss}

Stage~I optimizes a multi-task objective that jointly supervises appliance-level power estimation and activation state detection. The power estimation branch is trained using mean squared error (MSE) on the z-normalized appliance power, while the activation branch uses binary cross-entropy (BCE) on binary on/off labels. Normalizing the power target aligns the scale of the regression loss with that of the classification loss, allowing both tasks to contribute comparably. As a result, each loss is assigned equal weight, and the total training loss is defined as the sum of the two terms:

\begin{equation}\label{eq:loss_function}
\mathcal{L} = \mathcal{L}_{\mathrm{MSE}} + \mathcal{L}_{\mathrm{BCE}},
\end{equation}

\subsection{Stage II: Target Domain Embedding Learning}
\label{sec:stage2}

Stage~II adapts RefQuery to a target building through a lightweight appliance-specific adaptation stage. Starting from the model trained in Stage~I, the parameters of the shared feature extractor $f_{\theta}(\cdot)$ and the shared head $h_{\phi}(\cdot,\cdot)$ are kept fixed. Adaptation is performed by learning, incrementally, for each target appliance $k$, a target-specific reference embedding $\tilde{e}_r^{(k)} \in \mathbb{R}^{E}$, initialized at random. This makes the procedure lightweight because only $E$ parameters are optimized per appliance, while the full network remains unchanged. The embedding $\tilde{e}_r^{(k)}$ is optimized by minimizing the same loss function (\ref{eq:loss_function}). After adaptation, the learned embedding $\tilde{e}_r^{(k)}$ is fixed and reused for subsequent inference.

This design has three key practical benefits. It enables incremental on-site personalization without retraining the shared network, avoiding costly updates and preserving previously learned knowledge. This means that it can be easily scaled to new appliances by running the lightweight adaptation phase to learn the appliance fingerprint. It keeps the adaptation footprint predictable and extremely small (scaling only with $E$). Finally, because only a few parameters are optimized, the procedure is naturally compatible with limited label target data, reducing the risk of overfitting during adaptation while still capturing building-specific appliance characteristics.

\subsection{Stage III: Inference in the Target Domain}
\label{sec:stage3}

In Stage~III, RefQuery performs appliance disaggregation in the target building by reusing the reference embedding learned in Stage~II. Let $\tilde{e}_r^{(k)} \in \mathbb{R}^{E}$ denote the fixed target-domain reference embedding for appliance $k$. Given a sequence of successive query mains windows $q \in \mathbb{R}^{L}$ extracted from the target aggregate signal, each window is encoded by the frozen feature extractor as $e_q = f_{\theta}(q)$, and the frozen head is conditioned on $\tilde{e}_r^{(k)}$ to produce the appliance-specific outputs. For every query window, the model $h_{\phi}(\cdot,\cdot)$ returns the predicted activation state $\hat{s}^{(k)}$ and the predicted z-normalized power $\hat{y}^{(k)}$ of the appliance. Finally, $\hat{y}^{(k)}$ is mapped back to Watts by applying the inverse of the z-normalization parameters fitted in Stage~I, yielding appliance-level power estimates.

\section{Architecture}
\label{sec:arch}

\subsection{Lightweight convolutional encoder and embedding generation}
The RefQuery encoder $f_{\theta}(\cdot)$ is a compact 1-D CNN shared by both the reference and query branches, during the training phase in stage I, whose structure is selected to satisfy typical edge-device constraints on memory footprint, latency, and compute. In particular, we adopt a shallow stack of five convolutional layers with small kernels ($k=3$), ReLU activations, and same padding, interleaved with max-pooling to progressively reduce the temporal resolution while increasing the number of channels. This configuration limits parameter count and intermediate activation sizes while retaining the ability to capture the local temporal motifs and step-like transitions that characterize appliance activations in aggregate power signals, consistent with established CNN-based NILM backbones \cite{Kong2020,Zhou2024}.

\begin{figure*}[!t]
    \centering
    \includegraphics[width=1\linewidth]{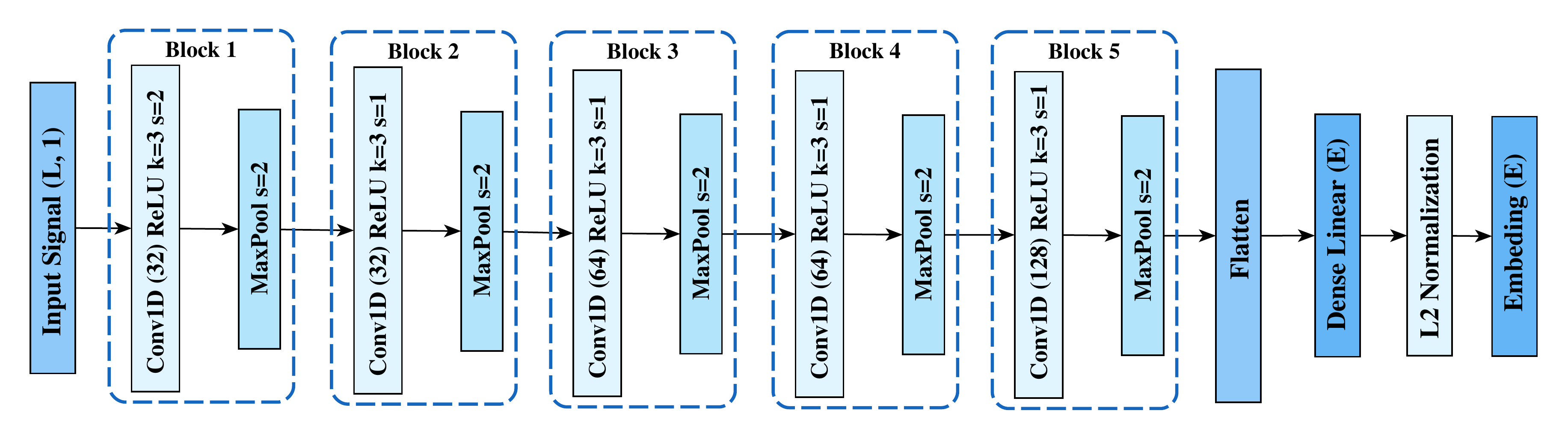}
    \caption{Compact 1D CNN feature extractor for embedding generation.}

    \label{fig:feature_extractor}
\end{figure*}

To reduce inference cost, downsampling is applied at the earliest stage of the encoder. The first convolutional layer uses a stride of 2 and is immediately followed by a stride-2 max-pooling layer, reducing the sequence length within the first block. This early temporal compression reduces the size of intermediate feature maps and limits memory movement, both of which significantly affect latency and energy consumption on resource-constrained hardware. After the last pooling stage, the features are flattened and linearly projected to an $E$-dimensional embedding. Finally, we apply L2 normalization to control embedding scale and improve the stability of the conditioning/matching operations used downstream, as commonly adopted in embedding-based formulations. The full encoder architecture is shown in Figure~\ref{fig:feature_extractor}.

\subsection{Multitask prediction head}
Given the reference and query embeddings $\mathbf{e}_r,\mathbf{e}_q \in \mathbb{R}^{E}$, the head builds a fused matching representation by concatenating the two embeddings with two lightweight interaction terms: the element-wise squared difference $(\mathbf{e}_q-\mathbf{e}_r)^2$ and the element-wise product $\mathbf{e}_q\odot\mathbf{e}_r$. The squared-difference term emphasizes coordinate-wise mismatch, while the product highlights shared patterns; both are commonly used in metric-based matching models to improve discrimination with minimal additional compute. The resulting $4E$ dimensional vector passes through two fully connected layers with RELU activation and width $H$, producing a shared latent representation that both tasks reuse.

\begin{figure*}[!t]
    \centering
    \includegraphics[width=1\linewidth]{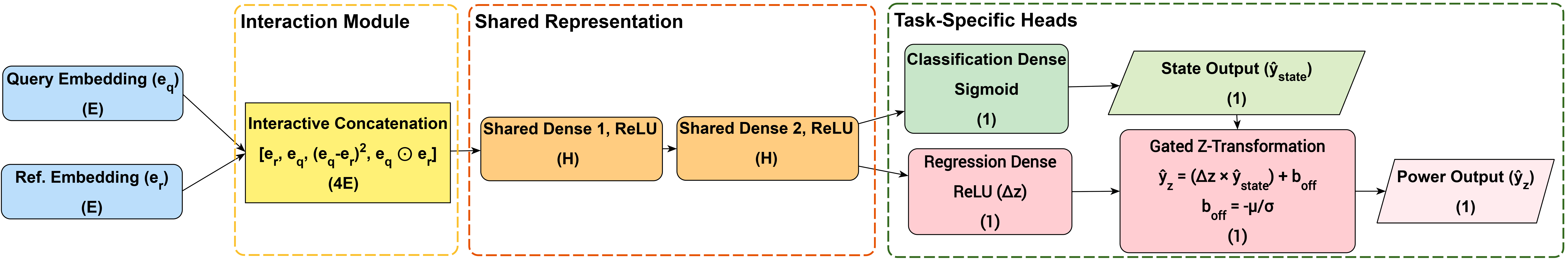}
    \caption{Multitask head architecture. Given the reference and query embeddings $\mathbf{e}_r,\mathbf{e}_q\in\mathbb{R}^{E}$, the interaction module constructs a joint representation by concatenating $[\mathbf{e}_r,\; \mathbf{e}_q,\; (\mathbf{e}_q-\mathbf{e}_r)^2,\; \mathbf{e}_q \odot \mathbf{e}_r]$, where $\odot$ denotes element-wise multiplication. The resulting $4E$-dimensional vector is processed by two shared fully connected layers with ReLU activations and width $H$ to produce a shared representation. Two task-specific branches are then applied: a classification head (dense + sigmoid) predicts the state probability $\hat{y}_{\text{state}}\in\mathbb{R}$, and a regression head (dense + ReLU) estimates a non-negative magnitude $\Delta z\in\mathbb{R}$. The final standardized power estimate is computed through a gated transformation $\hat{y}_z = \Delta z\,\hat{y}_{\text{state}} + b_{\text{off}}$, where $b_{\text{off}}=-\mu/\sigma$ anchors the OFF operating point in $z$-space.}
    \label{fig:head}
\end{figure*}

RefQuery operates in a seq2point setting, the head outputs a single prediction per input window. The activation branch predicts the on/off state via a sigmoid unit. The power branch predicts the z-normalized appliance power coupled to the activation output through a gating mechanism. To anchor the OFF operating point, we add a fixed bias corresponding to zero Watts in z-space (computed from the fitted normalizer) using a frozen linear layer with unit gain and constant bias. This design yields a physically consistent head in which the state output directly modulates the power output while keeping the OFF baseline stable. The full multitask head architecture is shown in Figure~\ref{fig:head}.

\section{Experiments}
\label{sec:experiments}

\subsection{Experimental Setup}
\label{subsec:exp_setup}

\textbf{Datasets and domain split.}
We evaluate RefQuery on three widely used public NILM datasets: REDD \cite{kolter2011redd}, UK-DALE \cite{Kelly2015}, and REFIT \cite{Murray2017}. We consider REFIT as the source domain and REDD (Houses 1 and 3) and UK-DALE (Houses 1 and 2) as target domains. This choice follows established cross-dataset evaluation practices in NILM \cite{DIncecco2020}. Training was performed on all 20 REFIT houses using a shared interval (2014-06-01–2014-07-01) because it provides a continuous one-month segment of available data across all buildings.

\textbf{Appliances.}
We consider five target appliances: fridge (F), dishwasher (DW), washing machine (WM), kettle (K), and microwave (MW). These appliances are widely used in NILM benchmarks and transfer learning studies. They account for a substantial share of household electricity consumption in the considered datasets, making them practically relevant targets for disaggregation \cite{Murray2017,DIncecco2020}. As in the REDD dataset, the kettle is not present, we consider only F, DW, WM, and MW for this dataset.

\textbf{Target-Domain Adaptation and Testing Split.}
We refer to UK-DALE House~1 and 2 as UKDALE-1 and UKDALE-2, and to REDD House~1 and 3 as REDD-1 and REDD-3. The target-domain time ranges are as follows: UKDALE-1 (2013-06-05 to 2013-07-05), UKDALE-2 (2013-06-02 to 2013-07-02), REDD-1 (2011-04-18 to 2011-05-25), and REDD-3 (2011-04-18 to 2011-05-25). For each case, we design two scenarios. In the first scenario, we use 7 days for adaptation and the remaining days for testing. In the second scenario, we simulate a limited data setting by using only 1 day for adaptation and the remaining days for testing. We select the time ranges to ensure that each scenario includes an interval in which all target appliances exhibit activations during both the adaptation and testing periods.

\textbf{Sampling period and window length.}
All signals are processed at a sampling period of $s=8$\,s to match the native resolution of the source domain (REFIT) \cite{Murray2017}. Since REDD and UK-DALE provide measurements at different native rates, resampling is required for both target datasets to obtain a consistent 8\,s sampling period before adaptation \cite{kolter2011redd,Kelly2015}. We use a fixed window length of $L=599$ samples, consistent with seq2point-style configurations adopted in low-rate disaggregation and transfer settings \cite{DIncecco2020}. Here, an odd length is selected to preserve symmetry around the central point.

\textbf{Implementation Details.}
In Stage~I, the 20 buildings are randomly split into training and validation sets using an 80/20 partition. In consequence, 16 buildings are used for training, while 4 different buildings are used for validation. The model is trained using the Adam optimizer with a learning rate of $10^{-3}$ and a batch size of 1024. Early stopping is applied based on the validation loss, with a patience of 5 epochs. In Stage~II, the target reference embedding is initialized randomly and optimized using Adam with a learning rate of $10^{-2}$. This higher learning rate is suitable because adaptation involves updating only the $E$-dimensional embedding, while the rest of the network remains fixed, allowing for fast convergence. Early stopping is based on the training loss with a patience of 5 epochs, as adaptation is performed on limited target-domain data without a separate validation set.

\textbf{Metrics.}
Let $y^{(k)}[t]$ and $\hat{y}^{(k)}[t]$ denote the ground-truth and predicted power (in Watts) of appliance $k$ at sample $t$, over an evaluation interval of length $T$. We report:

\textbf{Mean Absolute Error (MAE):}
\begin{equation}
\mathrm{MAE}^{(k)} = \frac{1}{T}\sum_{t=1}^{T}\left|y^{(k)}[t]-\hat{y}^{(k)}[t]\right|.
\end{equation}

\textbf{F-measure (F1) for activation detection:}
from the predicted activation probability $\hat{s}^{(k)}[t]\in[0,1]$, we obtain a binary prediction $\tilde{s}^{(k)}[t]=\mathbb{1}\{\hat{s}^{(k)}[t]\ge 0.5\}$ and compute precision and recall from the resulting counts of $\mathrm{TP}$, $\mathrm{FP}$, and $\mathrm{FN}$. The F1-score is 
\begin{equation}
\mathrm{F1}^{(k)} = \frac{2\,\mathrm{TP}}{2\,\mathrm{TP}+\mathrm{FP}+\mathrm{FN}}.
\end{equation}

Here, TP (true positive) denotes the number of instances in which an appliance is correctly detected as ON, FP (false positive) represents the number of instances in which an appliance is incorrectly detected as ON, and FN (false negative) corresponds to the number of instances in which an appliance is incorrectly detected as OFF.

\subsection{Baselines}
\label{subsec:baselines}
We compare RefQuery against three transfer-oriented NILM baselines:

\textbf{SA-S2P (D'Incecco et al.)} \cite{DIncecco2020}: a transfer-learning NILM baseline that follows the classical one-model-per-appliance paradigm based on a seq2point CNN.
As proposed in \cite{DIncecco2020}, cross-domain transfer is performed by keeping the convolutional feature extractor fixed and fine-tuning only the final dense layers. This baseline provides a classic but strong single-appliance reference.

\textbf{MA-CNN (Li et al.)} \cite{Li2023}: a transfer-learning NILM framework that replaces the traditional one-model-per-appliance paradigm with a compact one-to-many CNN. As proposed in the paper, cross-domain adaptation is performed via supervised retraining of the top dense layers and subsequent fine-tuning of the shared CNN layers. The method is explicitly motivated by reduced storage and computation.

\textbf{MA-TRF (Sun et al.)} \cite{Sun2025}: a recent multi-appliance transfer-learning NILM framework that combines attention mechanisms and Transformer blocks with a subtask-gated regression-classification design. As proposed in \cite{Sun2025}, transfer learning is performed by freezing the shared module and adapting to the target domain by fine-tuning only the appliance-specific layers. This baseline is designed to improve post-transfer accuracy, at the cost of higher model complexity than CNN-based alternatives.

Since SA-S2P and MA-CNN do not include a state detection module, we determine appliance states by comparing the model output to an activation power threshold value. We compute a threshold for each appliance by selecting the value that maximizes the F1 score on the adaptation set.

\subsection{Sensitivity analysis}
\label{subsec:sensitivity_analysis}

We first study the sensitivity of RefQuery to the embedding dimension $E$ and the width $H$ of the shared MLP in the multitask head. These parameters directly control the representational capacity of the reference-query matching function and, at the same time, the computational and memory footprint of the model. We evaluate a logarithmic grid $E,H \in \{32,64,128,256\}$, spanning compact to higher-capacity configurations while keeping all other settings fixed.

All configurations are trained in REFIT following Stage~I. Adaptation is performed using one-week of REDD-1 data, and performance is evaluated on the remaining data of the same dataset. We select REDD-1 because it represents a more challenging transfer scenario, since REFIT is collected in the UK, whereas REDD is collected in the US, leading to a larger domain shift.

\begin{table}[ht]
\caption{Top five hyperparameter configurations ranked by mean absolute error (MAE) averaged across appliances. Per-appliance MAE is also reported.}
\label{tab:top5_mae_appliances}
\centering
\begin{adjustbox}{width=\linewidth}
\begin{tabular}{r rr@{\hspace{10pt}}rrrr@{\hspace{10pt}}r}
\toprule
\multicolumn{1}{c}{} & \multicolumn{2}{c}{Config} & \multicolumn{4}{c}{MAE per appliance} & \multicolumn{1}{c}{} \\
\cmidrule(lr){2-3}\cmidrule(lr){4-7}
Rank & $E$ & $H$ & F & DW & MW & WM & Avg.\ MAE \\
\midrule
1 & 128 & 128 & 20.05 & 7.08 & 15.26 & 18.33 & 15.18 \\
2 & 64 & 256 & 19.01 & 10.53 & 14.35 & 19.56 & 15.86 \\
3 & 256 & 128 & 20.23 & 10.50 & 15.06 & 18.45 & 16.06 \\
4 & 256 & 32 & 19.56 & 9.40 & 15.33 & 20.09 & 16.09 \\
5 & 256 & 64 & 18.84 & 10.86 & 15.93 & 18.80 & 16.11 \\
\bottomrule
\end{tabular}
\end{adjustbox}
\end{table}

Table~\ref{tab:top5_mae_appliances} reports the top hyperparameter configurations ranked by average MAE.
The best-performing setting is the balanced configuration $(E,H)=(128,128)$, which reduces the average MAE by about 4\% relative to the second-ranked configuration. Table~\ref{tab:top5_mae_appliances} further shows that enlarging the embedding to $E=256$ does not improve the ranking among the top candidates. This pattern suggests diminishing returns from scaling the embedding under domain shift, where additional parameters may not translate into more transferable reference-query matching. At the appliance level, the ranking is primarily driven by DW and WM, whose errors vary more substantially across configurations than MW. Consequently, the $(128,128)$ setting is adopted for the remaining experiments.

\subsection{Comparison with the baselines}
\label{subsec:disc_baselines}

This section reports a comparison against the baselines for the first scenario, using seven days of data for adaptation. Table~\ref{tab:ukdale_results} shows that RefQuery obtains the best overall regression performance on UKDALE-1, yielding the lowest average MAE and improving over the Transformer-based MA-TRF by approximately 2.5\%. Moreover, RefQuery consistently outperforms both SA-S2P and MA-CNN in this house, confirming that the proposed approach improves upon the classical single-appliance transfer pipeline as well as compact multi-appliance CNN transfer baselines.

On UKDALE-2, Table~\ref{tab:ukdale_results} shows that RefQuery attains the second-best average MAE, with only a small gap to the top-performing MA-TRF, while clearly outperforming SA-S2P and MA-CNN.
At the same time, RefQuery achieves the best average F1, improving over MA-TRF by roughly 4\%. Considering both houses, Table~\ref{tab:ukdale_results} suggests that RefQuery shows the most consistent gains on multi-state appliances (DW and WM), consistently ranking first, with only one isolated case in which it ranks second. Table~\ref{tab:ukdale_results} also reveals that RefQuery MW detection can be unstable in some settings, obtaining a low F1 despite a competitive MAE. This divergence is characteristic of sparse, short activations typical in MW, for which event-level metrics are highly sensitive.

\begin{table}[t]
\centering
\caption{Performance comparison on UK-DALE.}
\label{tab:ukdale_results}
\renewcommand{\arraystretch}{1.10}
\setlength{\tabcolsep}{3pt}

\begin{adjustbox}{width=\linewidth}
\footnotesize

\begin{tabular}{l l | r r r r r r >{\centering\arraybackslash}p{0.80cm}@{}}
\toprule
\multirow{2}{*}{Metric} & \multirow{2}{*}{Method} & \multicolumn{6}{c}{UK-DALE} & \multirow{2}{0.80cm}{\centering House} \\
\cline{3-8}
 & & \multicolumn{1}{c}{F} & \multicolumn{1}{c}{DW} & \multicolumn{1}{c}{MW} & \multicolumn{1}{c}{WM} & \multicolumn{1}{c}{K} & \multicolumn{1}{c}{Ave} & \\
\midrule
\multirow{8}{*}{MAE\ $\downarrow$} & SA-S2P & 15.54 & 4.41 & 7.49 & 8.21 & 5.58 & 8.25 & \multirow{4}{*}{1} \\
 & MA-CNN & 21.15 & 4.83 & 7.25 & 8.35 & 8.50 & 10.01 &  \\
 & MA-TRF & \textbf{12.32} & 2.62 & \textbf{5.12} & 14.89 & \textbf{2.74} & 7.54 &  \\
 & RefQuery & 16.79 & \textbf{2.15} & 5.52 & \textbf{6.78} & 5.49 & \textbf{7.35} &  \\
\cline{2-8}\cline{9-9}
 & SA-S2P & 9.57 & 9.45 & 5.56 & 8.08 & 6.32 & 7.80 & \multirow{4}{*}{2} \\
 & MA-CNN & 10.62 & 11.39 & 10.91 & 12.25 & 10.95 & 11.22 &  \\
 & MA-TRF & 8.80 & \textbf{4.66} & \textbf{2.99} & 6.47 & \textbf{2.73} & \textbf{5.13} &  \\
 & RefQuery & \textbf{8.51} & 7.99 & 4.93 & \textbf{4.88} & 6.41 & 6.54 &  \\
\midrule
\multirow{8}{*}{F1\ $\uparrow$} & SA-S2P & 0.874 & 0.752 & 0.168 & 0.778 & 0.674 & 0.649 & \multirow{4}{*}{1} \\
 & MA-CNN & 0.827 & 0.678 & 0.224 & 0.745 & 0.377 & 0.570 &  \\
 & MA-TRF & \textbf{0.896} & 0.830 & \textbf{0.383} & 0.559 & \textbf{0.921} & \textbf{0.718} &  \\
 & RefQuery & 0.880 & \textbf{0.904} & 0.010 & \textbf{0.866} & 0.753 & 0.682 &  \\
\cline{2-8}\cline{9-9}
 & SA-S2P & 0.916 & 0.801 & 0.247 & 0.533 & 0.910 & 0.682 & \multirow{4}{*}{2} \\
 & MA-CNN & 0.905 & 0.674 & 0.180 & 0.433 & 0.558 & 0.550 &  \\
 & MA-TRF & 0.915 & 0.781 & \textbf{0.292} & 0.588 & \textbf{0.973} & 0.710 &  \\
 & RefQuery & \textbf{0.939} & \textbf{0.844} & 0.204 & \textbf{0.836} & 0.868 & \textbf{0.738} &  \\
\bottomrule
\end{tabular}
\end{adjustbox}
\end{table}

Table~\ref{tab:redd_results} reports the results on REDD. On REDD-1, RefQuery achieves the best average F1. In terms of MAE, MA-TRF attains the lowest value, but RefQuery remains very close, trailing by only about 4--5\%. The F1 improvement of RefQuery is mainly associated with multi-state appliances. For DW, RefQuery increases the F1 score over MA-TRF by approximately 6\% while reducing the mean absolute error by nearly 19\%. A similar pattern appears for WM, where RefQuery raises the F1 score by about 4\% and results in only a modest mean absolute error increase of around 4\%.

On REDD-3, Table~\ref{tab:redd_results} shows that RefQuery obtains the lowest MAE on three appliances (F, DW, and MW). However, the average MAE is dominated by WM, where activations are unusually high (around 5~kW) in this house, so small relative mismatches translate into large absolute MAE, and MA-TRF is more stable on this appliance, which largely explains the gap in the overall average. In addition, DW and MW have a few activations in this house, making F1 highly sensitive and contributing to the lower detection scores on these appliances.

\begin{table}[t]
\centering
\caption{Performance comparison on REDD.}
\label{tab:redd_results}
\renewcommand{\arraystretch}{1.10}
\setlength{\tabcolsep}{3pt}

\begin{adjustbox}{width=\linewidth}
\footnotesize

\begin{tabular}{l l | r r r r r >{\centering\arraybackslash}p{0.80cm}@{}}
\toprule
\multirow{2}{*}{Metric} & \multirow{2}{*}{Method} & \multicolumn{5}{c}{REDD} & \multirow{2}{0.80cm}{\centering House} \\
\cline{3-7}
 & & \multicolumn{1}{c}{F} & \multicolumn{1}{c}{DW} & \multicolumn{1}{c}{MW} & \multicolumn{1}{c}{WM} & \multicolumn{1}{c}{Ave} & \\
\midrule
\multirow{8}{*}{MAE\ $\downarrow$} & SA-S2P & 19.25 & 11.98 & 14.36 & 28.10 & 18.42 & \multirow{4}{*}{1} \\
 & MA-CNN & 26.42 & 9.05 & 18.46 & 31.54 & 21.37 &  \\
 & MA-TRF & \textbf{17.57} & 8.70 & \textbf{13.33} & \textbf{17.53} & \textbf{14.28} &  \\
 & RefQuery & 19.53 & \textbf{7.08} & 14.95 & 18.20 & 14.94 &  \\
\cline{2-7}\cline{8-8}
 & SA-S2P & 26.18 & 8.52 & 12.47 & 75.44 & 30.65 & \multirow{4}{*}{3} \\
 & MA-CNN & 33.45 & 12.96 & 10.47 & 48.43 & 26.33 &  \\
 & MA-TRF & 22.62 & 7.48 & 9.10 & \textbf{14.91} & \textbf{13.53} &  \\
 & RefQuery & \textbf{22.31} & \textbf{6.55} & \textbf{9.09} & 70.86 & 27.20 &  \\
\midrule
\multirow{8}{*}{F1\ $\uparrow$} & SA-S2P & 0.810 & 0.715 & \textbf{0.518} & 0.532 & 0.644 & \multirow{4}{*}{1} \\
 & MA-CNN & 0.827 & 0.801 & 0.414 & 0.279 & 0.580 &  \\
 & MA-TRF & \textbf{0.869} & 0.818 & 0.224 & 0.661 & 0.643 &  \\
 & RefQuery & 0.866 & \textbf{0.869} & 0.182 & \textbf{0.688} & \textbf{0.651} &  \\
\cline{2-7}\cline{8-8}
 & SA-S2P & 0.800 & \textbf{0.441} & 0.343 & 0.694 & \textbf{0.569} & \multirow{4}{*}{3} \\
 & MA-CNN & 0.753 & 0.163 & 0.207 & 0.832 & 0.489 &  \\
 & MA-TRF & 0.825 & 0.000 & \textbf{0.423} & \textbf{0.929} & 0.544 &  \\
 & RefQuery & \textbf{0.843} & 0.080 & 0.080 & 0.670 & 0.418 &  \\
\bottomrule
\end{tabular}

\end{adjustbox}
\end{table}

\subsection{Performance in Limited Data Scenarios}
\label{subsec:disc_limited}

In the following, we evaluate the methods under a limited data scenario in which only one day of data from each building is used for adaptation. This setting results in very few activations per appliance. For example, for DW and WM, only a single activation occurs within the adaptation period. Table~\ref{tab:ukdale_results_ld} shows the UKDALE-1 results. Under this constraint, RefQuery becomes the strongest overall method, improving the average MAE by about 8\% relative to MA-TRF and widening its margin over SA-S2P and MA-CNN. In addition, RefQuery yields the highest average F1.

\begin{table}[t]
\centering
\caption{Performance comparison on UK-DALE Limited Data.}
\label{tab:ukdale_results_ld}
\renewcommand{\arraystretch}{1.10}
\setlength{\tabcolsep}{3pt}

\begin{adjustbox}{width=\linewidth}
\footnotesize

\begin{tabular}{l l | r r r r r r >{\centering\arraybackslash}p{0.80cm}@{}}
\toprule
\multirow{2}{*}{Metric} & \multirow{2}{*}{Method} & \multicolumn{6}{c}{UK-DALE} & \multirow{2}{0.80cm}{\centering House} \\
\cline{3-8}
 & & \multicolumn{1}{c}{F} & \multicolumn{1}{c}{DW} & \multicolumn{1}{c}{MW} & \multicolumn{1}{c}{WM} & \multicolumn{1}{c}{K} & \multicolumn{1}{c}{Ave} & \\
\midrule
\multirow{4}{*}{MAE\ $\downarrow$} & SA-S2P & 21.95 & 11.69 & 7.99 & 13.62 & 12.18 & 13.49 & \multirow{4}{*}{1} \\
 & MA-CNN & 32.21 & 11.23 & 10.08 & 16.37 & 15.45 & 17.07 &  \\
 & MA-TRF & \textbf{13.28} & 4.05 & \textbf{5.32} & 19.42 & \textbf{4.41} & 9.30 &  \\
 & RefQuery & 20.50 & \textbf{2.54} & 5.45 & \textbf{7.84} & 6.31 & \textbf{8.53} &  \\
\midrule
\multirow{4}{*}{F1\ $\uparrow$} & SA-S2P & 0.735 & 0.450 & 0.315 & 0.575 & 0.403 & 0.496 & \multirow{4}{*}{1} \\
 & MA-CNN & 0.729 & 0.262 & 0.098 & 0.462 & 0.192 & 0.349 &  \\
 & MA-TRF & \textbf{0.881} & 0.453 & \textbf{0.509} & 0.333 & \textbf{0.875} & 0.610 &  \\
 & RefQuery & 0.814 & \textbf{0.898} & 0.016 & \textbf{0.821} & 0.736 & \textbf{0.657} &  \\
\bottomrule
\end{tabular}

\end{adjustbox}
\end{table}

The robustness of the methods is quantified in Table~\ref{tab:ukdale_percent_changes_h1}, which reports relative degradations from 7-day to 1-day adaptation data on UKDALE-1. RefQuery shows the smallest deterioration, reducing the MAE degradation by about 31\% relative to MA-TRF, and shrinking the F1 drop by approximately 75\% relative to MA-TRF. These relative improvements are particularly relevant for practical deployment, where adaptation data are typically scarce and costly to obtain.

\begin{table}[t]
\centering
\caption{Relative changes on UK-DALE (Limited Data vs. Normal) for House 1, in \%. Positive $\Delta$MAE indicates worse error; negative $\Delta$F1 indicates worse accuracy. Best (per appliance/column) is in bold.}
\label{tab:ukdale_percent_changes_h1}
\renewcommand{\arraystretch}{1.10}
\setlength{\tabcolsep}{3pt}
\footnotesize 
\begin{tabular}{l l | r r r r r r @{}}
\toprule
\multirow{2}{*}{Metric} & \multirow{2}{*}{Method} & \multicolumn{6}{c}{UK-DALE (\% change)} \\
\cline{3-8}
 & & \multicolumn{1}{c}{F} & \multicolumn{1}{c}{DW} & \multicolumn{1}{c}{MW} & \multicolumn{1}{c}{WM} & \multicolumn{1}{c}{K} & \multicolumn{1}{c}{Ave} \\
\midrule
\multirow{4}{*}{$\Delta$MAE}
 & SA-S2P   & 41.2 & 165.1 & 6.7  & 65.9 & 118.3 & 63.5 \\
 & MA-CNN   & 52.3 & 132.5 & 39.0 & 96.0 & 81.8  & 70.5 \\
 & MA-TRF   & \textbf{7.8} & 54.6 & 3.9 & 30.4 & 60.9 & 23.3 \\
 & RefQuery & 22.1 & \textbf{18.1} & \textbf{-1.3} & \textbf{15.6} & \textbf{14.9} & \textbf{16.1} \\
\midrule
\multirow{4}{*}{$\Delta$F1}
 & SA-S2P   & -15.9 & -40.2 & \textbf{87.5} & -26.1 & -40.2 & -23.6 \\
 & MA-CNN   & -11.9 & -61.4 & -56.2 & -38.0 & -49.1 & -38.8 \\
 & MA-TRF   & \textbf{-1.7} & -45.4 & 32.9 & -40.4 & -5.0 & -15.0 \\
 & RefQuery & -7.5 & \textbf{-0.7} & 60.0 & \textbf{-5.2} & \textbf{-2.3} & \textbf{-3.7} \\
\bottomrule
\end{tabular}
\end{table}

Table~\ref{tab:redd_results_ld} reports limited-data performance on REDD-1. Here, MA-TRF retains the best MAE, but RefQuery remains within a single-digit percentage increase in MAE while improving average F1 by roughly 20\% relative to MA-TRF.
This pattern is reinforced by Table~\ref{tab:redd_percent_changes_h1}, where RefQuery reduces the F1 degradation by about 70\% compared with MA-TRF. Therefore, RefQuery consistently offers substantially higher stability in detection accuracy as supervision is reduced.

\begin{table}[t]
\centering
\caption{Performance comparison on REDD Limited Data.}
\label{tab:redd_results_ld}
\renewcommand{\arraystretch}{1.10}
\setlength{\tabcolsep}{3pt}

\begin{adjustbox}{width=\linewidth}
\footnotesize

\begin{tabular}{l l | r r r r r >{\centering\arraybackslash}p{0.80cm}@{}}
\toprule
\multirow{2}{*}{Metric} & \multirow{2}{*}{Method} & \multicolumn{5}{c}{REDD} & \multirow{2}{0.80cm}{\centering House} \\
\cline{3-7}
 & & \multicolumn{1}{c}{F} & \multicolumn{1}{c}{DW} & \multicolumn{1}{c}{MW} & \multicolumn{1}{c}{WM} & \multicolumn{1}{c}{Ave} & \\
\midrule
\multirow{4}{*}{MAE\ $\downarrow$} & SA-S2P & 25.16 & 17.65 & 16.16 & 30.44 & 22.35 & \multirow{4}{*}{1} \\
 & MA-CNN & 33.43 & 18.44 & 31.25 & 29.00 & 28.03 &  \\
 & MA-TRF & \textbf{20.23} & 9.29 & \textbf{15.94} & 20.91 & \textbf{16.59} &  \\
 & RefQuery & 22.69 & \textbf{7.96} & 20.69 & \textbf{19.75} & 17.77 &  \\
\midrule
\multirow{4}{*}{F1\ $\uparrow$} & SA-S2P & 0.750 & 0.627 & \textbf{0.426} & 0.262 & 0.516 & \multirow{4}{*}{1} \\
 & MA-CNN & 0.757 & 0.624 & 0.268 & 0.192 & 0.460 &  \\
 & MA-TRF & \textbf{0.855} & 0.736 & 0.000 & 0.440 & 0.508 &  \\
 & RefQuery & 0.835 & \textbf{0.856} & 0.175 & \textbf{0.579} & \textbf{0.611} &  \\
\bottomrule
\end{tabular}

\end{adjustbox}
\end{table}

\begin{table}[t]
\centering
\caption{Relative changes on REDD (Limited Data vs. Normal) for House 1, in \%. Positive $\Delta$MAE indicates worse error; negative $\Delta$F1 indicates worse accuracy. Best (per appliance/column) is in bold.}
\label{tab:redd_percent_changes_h1}
\renewcommand{\arraystretch}{1.0}
\setlength{\tabcolsep}{3pt}

\footnotesize 

\begin{tabular}{l l | r r r r r @{}}
\toprule
\multirow{2}{*}{Metric} & \multirow{2}{*}{Method} & \multicolumn{5}{c}{REDD House 1 (\% change)} \\
\cline{3-7}
 & & \multicolumn{1}{c}{F} & \multicolumn{1}{c}{DW} & \multicolumn{1}{c}{MW} & \multicolumn{1}{c}{WM} & \multicolumn{1}{c}{Ave} \\
\midrule
\multirow{4}{*}{$\Delta$MAE}
 & SA-S2P   & 30.7 & 47.3  & \textbf{12.5} & 8.3  & 21.3 \\
 & MA-CNN   & 26.5 & 103.8 & 69.3          & \textbf{-8.1} & 31.2 \\
 & MA-TRF   & \textbf{15.1} & \textbf{6.8} & 19.6 & 19.3 & \textbf{16.2} \\
 & RefQuery & 16.2 & 12.4  & 38.4          & 8.5  & 18.9 \\
\midrule
\multirow{4}{*}{$\Delta$F1}
 & SA-S2P   & -7.4  & -12.3 & -17.8 & -50.8 & -19.9 \\
 & MA-CNN   & -8.5  & -22.1 & -35.3 & -31.2 & -20.7 \\
 & MA-TRF   & \textbf{-1.6} & -10.0 & -100.0 & -33.4 & -21.0 \\
 & RefQuery & -3.6  & \textbf{-1.5} & \textbf{-3.8} & \textbf{-15.8} & \textbf{-6.1} \\
\bottomrule
\end{tabular}

\end{table}

\subsection{Computational Performance and Scalability}
\label{subsec:disc_efficiency}

Table~\ref{tab:comp_compare_1app} compares storage, adaptation time, and inference time for single-appliance adaptation using 7 days of data followed by one-month inference. Computational performance was assessed under a CPU-only constrained execution regime, in which the benchmarking process was restricted to two CPU cores via process affinity, and thread-level parallelism was limited to two threads. This configuration was adopted to provide a controlled, reproducible, and edge-oriented comparison of runtime performance. This comparative analysis emphasizes offline performance. Consequently, inference time depends on implementation choices, including batch size and whether a method uses a seq2seq or seq2point formulation. For example, MA-TRF employs a transformer architecture, which requires smaller batches to fit in memory. However, because it uses a seq2seq setup with a window stride of 120 \cite{Sun2025}, it evaluates fewer windows to cover one month of data compared to a seq2point method. Nevertheless, this analysis reflects how each method behaves under the optimal settings reported in its original paper.

Here, RefQuery reduces storage by about 90\% relative to MA-TRF, about 50\% relative to MA-CNN, and by about 99.6\% relative to SA-S2P. It also achieves dramatically faster adaptation, being about 70$\times$ faster than MA-TRF, about 78$\times$ faster than MA-CNN, and about 85$\times$ faster than SA-S2P. Inference is likewise consistently faster, with RefQuery being about 5.9$\times$ faster than MA-TRF, about 2.7$\times$ faster than MA-CNN, and about 9.2$\times$ faster than SA-S2P. These results demonstrate the computational advantages of RefQuery over the baselines for offline applications, particularly in adaptation time.

\newcolumntype{Y}{>{\centering\arraybackslash}X}

\begin{table}[t]
\centering
\caption{Computational comparison of the methods for single-appliance adaptation using 7 days of data, followed by one-month offline inference, evaluated under a CPU-only constrained execution regime (process affinity limited to two CPU cores and thread-level parallelism limited to two threads).}
\label{tab:comp_compare_1app}
\renewcommand{\arraystretch}{1.08}
\setlength{\tabcolsep}{4pt}
\footnotesize

\begin{tabularx}{\linewidth}{l r Y Y}
\toprule
Method & Storage (MB)$\downarrow$ & Adaptation (s)$\downarrow$ & Inference (s)$\downarrow$ \\
\midrule
SA-S2P   & 334.50          & 5400.3         & 391.07 \\
MA-CNN   & 2.38            & 4950.5         & 115.11 \\
MA-TRF   & 11.81           & 4429.5         & 248.86 \\
RefQuery & \textbf{1.18}   & \textbf{63.2}  & \textbf{42.41} \\
\bottomrule
\end{tabularx}

\end{table}

On the other hand, Table~\ref{tab:scalability} summarizes scalability with the number of appliances. This comparison focuses on the practical use of each method for real-time NILM on constrained edge devices. We report inference scaling in terms of MFLOPs (million floating-point operations) per processed mains window, which aligns with sequential, window-by-window disaggregation. The table highlights that SA-S2P scales poorly because each new appliance requires an additional full model, whereas MA-CNN, although memory-light per appliance, is not incremental in practice since extending the appliance set requires updating shared parameters and full retraining. MA-TRF, on the other hand, supports incremental extension via additional heads but incurs a substantially higher overhead, limiting its practical use in real-time applications. 

RefQuery achieves both minimal per-appliance overhead and incremental deployment. New appliances can be incorporated without updating the network parameters, requiring only an appliance embedding. Inference scaling further supports the practicality of the proposed method. The computational overhead from the shared feature extractor is minimal. Simultaneously, the low per-appliance cost, driven by calls to the multitask head, makes this approach well-suited for real-time, resource-constrained deployments.

\begin{table}[t]
\centering
\caption{Scalability with number of appliances $K$. $\Delta W$ is the additional storage per added appliance. “Incremental” means adding an appliance without updating shared parameters. Inference scaling is reported in MFLOPs/(mains window) as $b + K\cdot m$, where $b$ is the $K$-independent compute (shared cost) and $m$ is the additional MFLOPs per added appliance.}
\label{tab:scalability}
\footnotesize
\renewcommand{\arraystretch}{1.08}
\setlength{\tabcolsep}{3pt}

\begin{tabularx}{\columnwidth}{@{}l c c X@{}}
\toprule
Method & $\Delta W$ (KB)$\downarrow$ & Incremental? & Inference scaling (MFLOPs)$\downarrow$ \\
\midrule
SA-S2P   & $3.4\times10^{5}$    & Yes$^{a}$ & $0.000 + K \cdot 101.289$ \\
MA-CNN   & \textbf{$\approx 1$} & No$^{c}$  & $13.295 + K \cdot 0.000129$ \\
MA-TRF   & $4.6\times10^{3}$    & Yes$^{b}$ & $889.560 + K \cdot 1142.726$ \\
RefQuery & \textbf{$\approx 1$} & Yes       & $4.145 + K \cdot 0.165$ \\
\bottomrule
\end{tabularx}

\vspace{0.6ex}
{\scriptsize $^{a}$ new model per appliance;\;
$^{b}$ add a Transformer head per appliance (backbone frozen);\;
$^{c}$ update shared multi-output mapping.}
\end{table}

\section{Conclusion}
\label{sec:conclusion}

This work proposed RefQuery, a transfer-oriented NILM framework designed for constrained edge deployment and scalable multi-appliance monitoring.
By conditioning disaggregation on compact appliance fingerprints, RefQuery departs from both one-model-per-appliance and fixed-set multi-output paradigms, enabling incremental extension of appliances without updating network parameters and with negligible per-appliance storage. A novel source-domain training strategy and target-domain adaptation mechanism are designed to obtain the appliance fingerprints through a lightweight training phase.

Experiments on three public datasets demonstrate that RefQuery delivers a strong accuracy-efficiency trade-off against single-appliance and multi-appliance baselines, including modern Transformer-based methods. Under limited data supervision, RefQuery becomes consistently preferable to the baselines. It preserves a competitive mean absolute error while delivering substantially higher F1 stability. Finally, RefQuery provides decisive deployment benefits, reducing model size and adaptation time, and offering a practical path toward scalable real-time NILM on constrained edge devices.


\bibliography{references}
\bibliographystyle{IEEEtran}


 





\end{document}